\documentclass[11pt,a4paper]{article}
\usepackage[hyperref]{acl2018}
\usepackage{times} 
\usepackage{latexsym}
\usepackage{amsmath} 
\usepackage{graphicx}
\usepackage{multirow}
\usepackage{subcaption}
\usepackage{array}

\usepackage{url}

\aclfinalcopy 
\def\aclpaperid{54} 

\title{Variational Cross-domain Natural Language Generation for Spoken Dialogue Systems}

\author{Bo-Hsiang Tseng, Florian Kreyssig, Pawe\l\ Budzianowski, \\\textbf{I\~nigo Casanueva, Yen-Chen Wu, Stefan Ultes, Milica~Ga{\v s}i{\' c}} \\
Department of Engineering, University of Cambridge, Cambridge, UK\\
   {\tt \{bht26,flk24,pfb30,ic340,ycw30,su259,mg436\}@cam.ac.uk} 
\\}

\date{}

\begin{document}
\maketitle
\begin{abstract}
Cross-domain natural language generation (NLG) is still a difficult task within spoken dialogue modelling. Given a semantic representation provided by the dialogue manager, the language generator should generate sentences that convey desired information. Traditional template-based generators can produce sentences with all necessary information, but these sentences are not sufficiently diverse. With RNN-based models, the diversity of the generated sentences can be high, however, in the process some information is lost. In this work, we improve an RNN-based generator by considering latent information at the sentence level during generation using the conditional variational autoencoder architecture. We demonstrate that our model outperforms the original RNN-based generator, while yielding highly diverse sentences. In addition, our model performs better when the training data is limited.
\end{abstract}

\section{Introduction}
Conventional spoken dialogue systems (SDS) require a substantial amount of hand-crafted rules to achieve good interaction with users. 
The large amount of required engineering limits the scalability of these systems to settings with new or multiple domains. Recently, statistical approaches have been studied that allow natural, efficient and more diverse interaction with users without depending on pre-defined rules~\citep{young2013pomdp,gavsic2014incremental,henderson2014robust}. 

Natural language generation (NLG) is an essential component of an SDS. Given a semantic representation (SR) consisting of a dialogue act and a set of slot-value pairs, the generator should produce natural language containing the desired information. 

Traditionally NLG was based on templates \citep{cheyer2014method}, which produce grammatically-correct sentences that contain all desired information. However, the lack of variation of these sentences made these systems seem tedious and monotonic. 
\textit{Trainable generators} \citep{langkilde1998generation,stent2004trainable} can generate several sentences for the same SR, but the dependence on pre-defined operations limits their potential. Corpus-based approaches \citep{oh2000stochastic,mairesse2011controlling} learn to generate natural language directly from data without pre-defined rules. However, they usually require alignment between the sentence and the SR. 
Recently, Wen et al.~\shortcite{wensclstm15} proposed an RNN-based approach, which outperformed previous methods on several metrics. However, the generated sentences often did not include all desired attributes.

The variational autoencoder~\citep{journals/corr/KingmaW13} enabled for the first time the generation of complicated, high-dimensional data such as images.
The conditional variational autoencoder (CVAE)~\citep{sohn2015learning}, firstly proposed for image generation, has a similar structure to the VAE with an additional dependency on a condition. 
Recently, the CVAE has been applied to dialogue systems \citep{serban2017hierarchical,Shen2017ACV,ZhaoZE17} using the previous dialogue turns as the condition. However, their output was not required to contain specific information.

In this paper, we improve RNN-based generators by adapting the CVAE to the difficult task of cross-domain NLG. 
Due to the additional latent information encoded by the CVAE, our model outperformed the SCLSTM at conveying all information. Furthermore, our model reaches better results when the training data is limited.

\section{Model Description}
\subsection{Variational Autoencoder}
The VAE is a generative latent variable model. It uses a neural network (NN) to generate $\hat{x}$ from a latent variable $z$, which is sampled from the prior $p_{\theta}(z)$. The VAE is trained such that $\hat{x}$ is a sample of the distribution $p_{D}(x)$ from which the training data was collected. Generative latent variable models have the form $p_{\theta}(x)=\int_{z}p_{\theta}(x|z)p_{\theta}(z) dz$. In a VAE an NN, called the decoder, models $p_{\theta}(x|z)$ and would ideally be trained to maximize the expectation of the above integral $E\left[p_{\theta}(x)\right]$. 
Since this is intractable, the VAE uses another NN, called the encoder, to model $q_{\phi}(z|x)$ which should approximate the posterior $p_{\theta}(z|x)$. The NNs in the VAE are trained to maximise the variational lower bound (VLB) to $\log p_{\theta}(x)$, which is given by:
\begin{equation}
\begin{aligned}
	L_{VAE}(\theta, \phi; x) = -KL(q_{\phi}(z|x)||p_{\theta}(z)) \\
    + E_{q_{\phi}(z|x)}[\log p_{\theta}(x|z)]
\label{eq:vae}
\end{aligned}
\end{equation}
The first term is the KL-divergence between the approximated posterior and the prior, which encourages similarity between the two distributions. The second term is the likelihood of the data given samples from the approximated posterior. The CVAE has a similar structure, but the prior is modelled by another NN, called the prior network. The prior network is conditioned on $c$. The new objective function can now be written as:
\begin{multline}
	L_{CVAE}(\theta, \phi; x, c) = -KL(q_{\phi}(z|x, c)||p_{\theta}(z|c)) \\
    + E_{q_{\phi}(z|x,c)}[\log p_{\theta}(x|z,c)]
\end{multline}
When generating data, the encoder is not used and $z$ is sampled from $p_{\theta}(z|c)$.

\subsection{Semantically Conditioned VAE}

\begin{figure}
	\includegraphics[width=\linewidth]{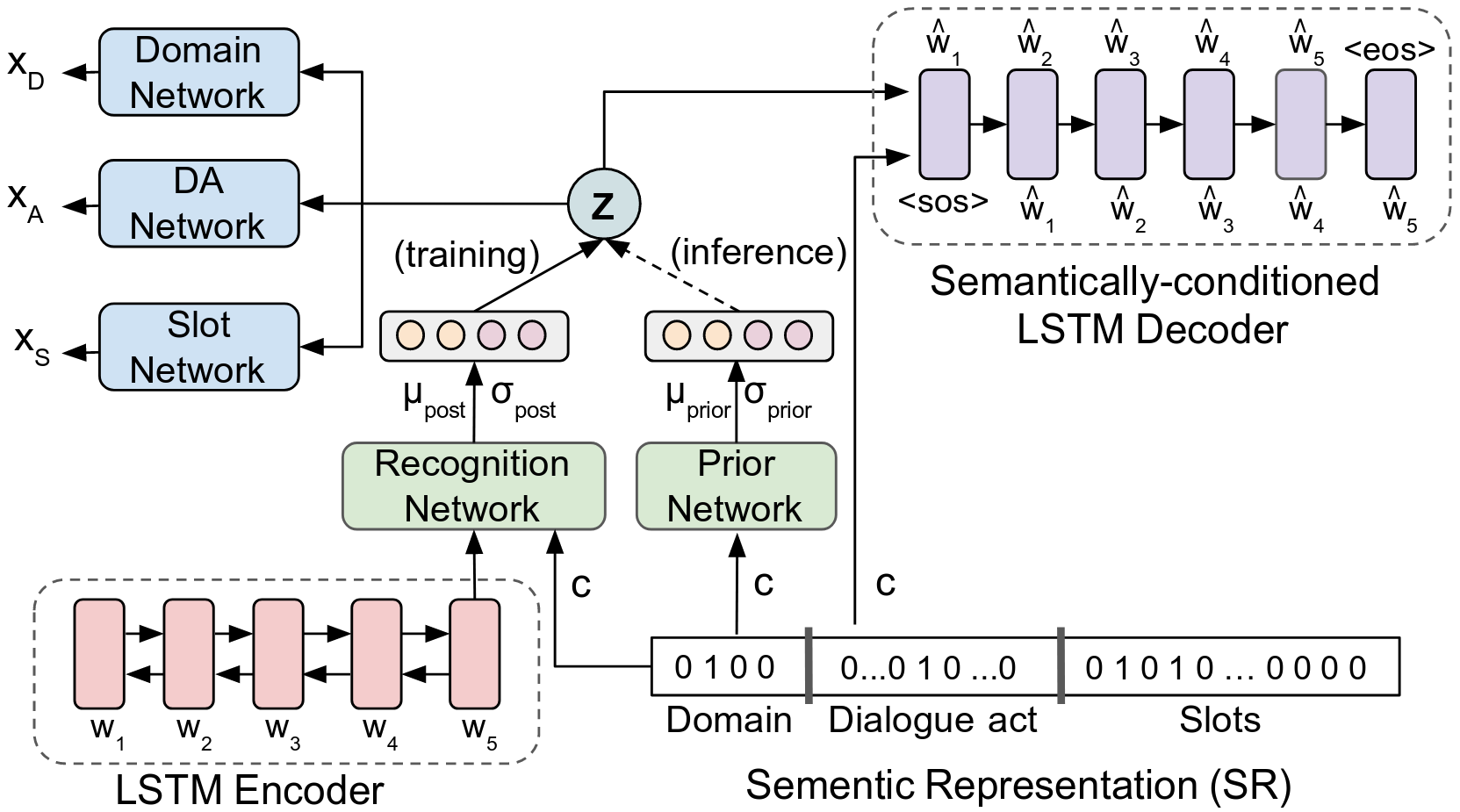}
    \caption{Semantically Conditioned Variational Autoencoder with a semantic representation (SR) as the condition. $x$ is the system response with words $w_{1:N}$. $x_{D}$, $x_{A}$ and $x_{S}$ are labels for the domain, the dialogue act (DA) and the slots of $x$.}
    \label{fig:scvae}
    \vspace{-0.5em}
\end{figure}


The structure of our model is depicted in Fig.~\ref{fig:scvae}, which, conditioned on an SR, generates the system's word-level response $x$. An SR consists of three components: the domain, a dialogue act and a set of slot-value pairs. \textit{Slots} are attributes required to appear in $x$ (e.g. a hotel's \textit{area}). A \textit{slot} can have a \textit{value}. Then the two are called a \textit{slot-value} pair (e.g. \textit{area}=\textit{north}). $x$ is \textit{delexicalised}, which means that slot values are replaced by corresponding slot tokens. The condition $c$ of our model is the SR represented as two 1-hot vectors for the domain and the dialogue act as well as a binary vector for the slots.

During training, $x$ is first passed through a single layer bi-directional LSTM, the output of which is concatenated with $c$ and passed to the recognition network. The recognition network parametrises a Gaussian distribution $\mathcal{N}(\mu_{post}, \sigma_{post})$ which is the posterior.The prior network only has $c$ as its input and parametrises a Gaussian distribution $\mathcal{N}(\mu_{prior}, \sigma_{prior})$ which is the prior. Both networks are fully-connected (FC) NNs with one and two layers respectively. During training, $z$ is sampled from the posterior. When the model is used for generation, $z$ is sampled from the prior. The decoder is an SCLSTM~\citep{wensclstm15} using $z$ as its initial hidden state and initial cell vector. The first input to the SCLSTM is a start-of-sentence (sos) token and the model generates words until it outputs an end-of-sentence (eos) token.

\subsection{Optimization}
When the decoder in the CVAE is powerful on its own, it tends to ignore the latent variable $z$ since the encoder fails to encode enough information into $z$. Regularization methods can be introduced in order to push the encoder towards learning a good representation of the latent variable $z$. 
Since the KL-component of the VLB does not contribute towards learning a meaningful $z$, increasing the weight of it gradually from $0$ to $1$ during training helps to encode a better representation in $z$. This method is termed \textit{KL-annealing} \citep{476cadd89dec4e0ab01d9dc59e1222c7}.
In addition, inspired by \citep{ZhaoZE17}, we introduce a regularization method using another NN which is trained to use $z$ to recover the condition $c$. The NN is split into three separate FC NNs of one layer each, which independently recover the \textit{domain}, \textit{dialogue-act} and \textit{slots} components of $c$. The objective of our model can be written as:
\begin{multline}
	L_{SCVAE}(\theta, \phi;x, c) = L_{CVAE}(\theta, \phi;x, c) \\
	+ E_{q_{\phi}(z|x, c)}[\log p(x_{D}|z)+\log p(x_{A}|z)+ \\
    \log \prod_{i=1}^{|S|} p(x_{S_{i}}|z)] 
\end{multline}
where $x_{D}$ is the domain label, $x_{A}$ is the dialogue act label and $x_{S_{i}}$ are the slot labels with $|S|$ slots in the SR.  
In the proposed model, the CVAE learns to encode information about both the sentence and the SR into $z$. 
Using $z$ as its initial state, the decoder is better at generating sentences with desired attributes. In section \ref{sec:z} a visualization of the latent space demonstrates that a semantically meaningful representation for $z$ was learned.


\begin{table*}[tp]
\centering
\caption{The statistics of the cross-domain dataset}
\label{tab:data}
\resizebox{\textwidth}{!}{%
\begin{tabular}{ccccc}
\hline
               & Restaurant                                                                  & Hotel                                                                               & Television                                                                                                                                                           & Laptop                                                                                                                                                                                     \\ \hline\hline
\# of examples & 3114/1039/1039                                                              & 3223/1075/1075                                                                      & 4221/1407/1407                                                                                                                                                       & 7944/2649/2649                                                                                                                                                                             \\ \hline
dialogue acts  & \multicolumn{2}{c}{\begin{tabular}[c]{@{}c@{}}reqmore, goodbye, select, confirm, request, \\ inform, inform\_only, inform\_count, inform\_no\_match\end{tabular}} & \multicolumn{2}{c}{\begin{tabular}[c]{@{}c@{}}compare, recommend, inform\_all, \\ suggest, inform\_no\_info, 9 acts as left\end{tabular}}                                                                                                                                                                                                                         \\ \hline
shared slots   & \multicolumn{2}{c}{\begin{tabular}[c]{@{}c@{}}name, type, area, near, price,\\ phone, address, postcode, pricerange\end{tabular}}                                 & \multicolumn{2}{c}{\begin{tabular}[c]{@{}c@{}}name, type, price,\\ family, pricerange,\end{tabular}}                                                                                                                                                                                                                                                              \\ \hline
specific slots & \begin{tabular}[c]{@{}c@{}}food,\\ goodformeal,\\ kids-allowed\end{tabular} & \begin{tabular}[c]{@{}c@{}}hasinternet,\\ acceptscards,\\ dogs-allowed\end{tabular} & \begin{tabular}[c]{@{}c@{}}screensizerange, ecorating, \\ hdmiport, hasusbport, audio,\\ accessories, color, screensize,\\ resolution, powerconsumption\end{tabular} & \begin{tabular}[c]{@{}c@{}}isforbusinesscomputing.\\ warranty, battery, design,\\ batteryrating, weightrange,\\ utility, platform, driverange,\\ dimension, memory, processor\end{tabular} \\ \hline
\end{tabular}%
}
\end{table*}

\section{Dataset and Setup}
The proposed model is used for an SDS that provides information about restaurants, hotels, televisions and laptops. It is trained on a dataset \cite{wenmultinlg16}, which consists of sentences with corresponding semantic representations.
Table~\ref{tab:data} shows statistics about the corpus which was split into a training, validation and testing set according to a 3:1:1 split.
The dataset contains 14 different system dialogue acts.
The television and laptop domains are much more complex than other domains. There are around 7k and 13k different SRs possible for the TV and the laptop domain respectively. For the restaurant and hotel domains only 248 and 164 unique SRs are possible.
This imbalance makes the NLG task more difficult.

The generators were implemented using the PyTorch Library~\citep{paszke2017automatic}.
The size of decoder SCLSTM and thus of the latent variable was set to 128.
KL-annealing was used, with the weight of the KL-loss reaching $1$ after 5k mini-batch updates.
The slot error rate (ERR), used in \citep{oh2000stochastic,thwsjy15}, is the metric that measures the model's ability to convey the desired information.
ERR is defined as: $(p+q)/N$, where $N$ is the number of slots in the SR, $p$ and $q$ are the number of missing and redundant slots in the generated sentence. 
The BLEU-4 metric and perplexity (PPL) are also reported. The baseline SCLSTM is optimized, which has shown to outperform template-based methods and trainable generators \cite{wensclstm15}.
NLG often uses the over-generation and reranking paradigm \cite{oh2000stochastic}. The SCVAE can generate multiple sentences by sampling multiple $z$, while the SCLSTM has to sample different words from the output distribution.
In our experiments ten sentences are generated per SR. Table~\ref{tab:example} in the appendix shows one SR in each domain with five illustrative sentences generated by our model.

\section{Experimental Results}
\subsection{Visualization of Latent Variable $z$} \label{sec:z}
\begin{figure}[tp]
\centering
	\includegraphics[width=\linewidth]{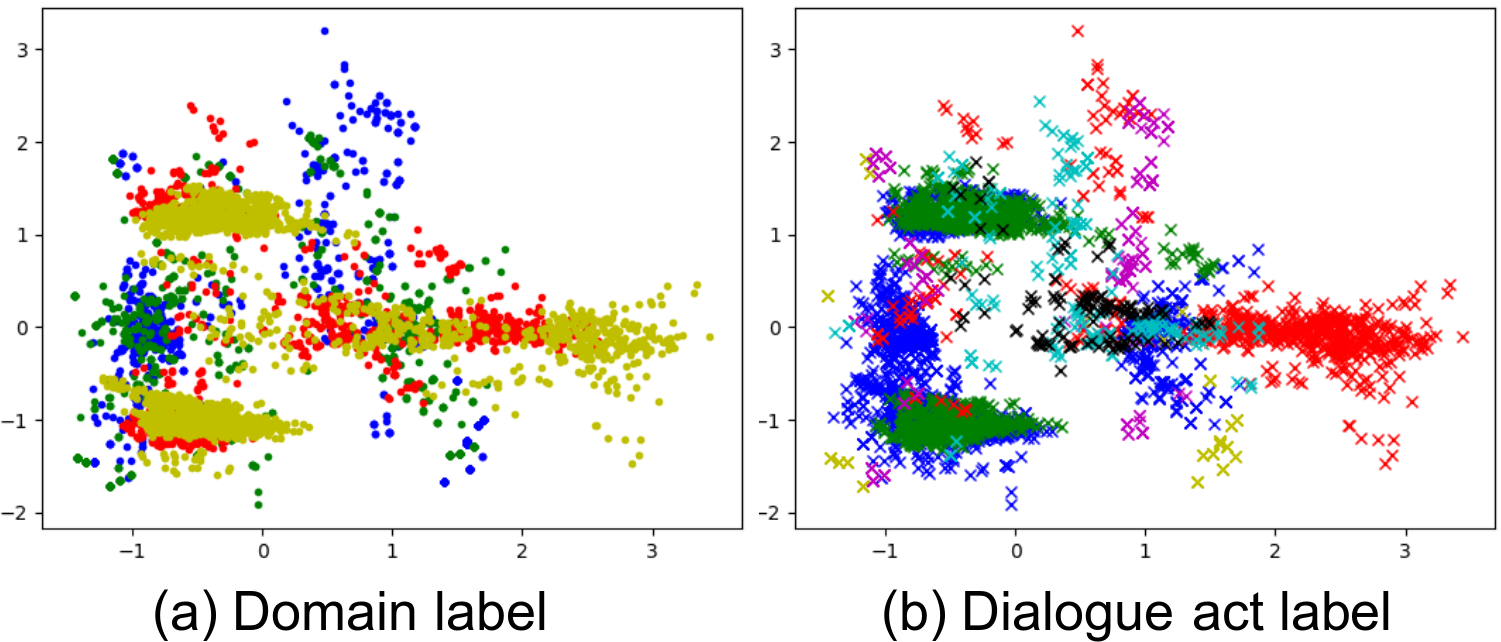}
\caption{2D-projection of $z$ for each data point in the test set, with two different colouring-schemes.}
\label{fig:pca}
\end{figure}
2D-projections of $z$ for each data point in the test set are shown in Fig.~\ref{fig:pca}, by using PCA for dimensionality reduction.
In Fig.~\ref{fig:pca}a, data points of the restaurant, hotel, TV and laptop domain are marked as blue, green, red and yellow respectively.
As can be seen, data points from the laptop domain are contained within four distinct clusters. In addition, 
there is a large overlap of the TV and laptop domains, which is not surprising as they share all dialogue acts (DAs).
Similarly, there is overlap of the restaurant and hotel domains.
In Fig.~\ref{fig:pca}b, the eight most frequent DAs are color-coded. \texttt{recommend}, depicted as green, has a similar distribution to the laptop domain in Fig.~\ref{fig:pca}a, since \texttt{recommend} happens mostly in the laptop domain.
This suggests that our model learns to map similar SRs into close regions within the latent space. Therefore, $z$ contains meaningful information in regards to the domain, DAs and slots.

\subsection{Empirical Comparison}


\begin{table}[tp]
\centering
\caption{Comparison between SCVAE and SCLSTM. Both are trained with full dataset and tested on individual domains}
\label{tab:indiv}
\resizebox{\columnwidth}{!}{%
\begin{tabular}{l|l|ccccc}
\hline
Metrics               & Method & Restaurant     & Hotel          & TV             & Laptop         & Overall        \\ \hline\hline
\multirow{2}{*}{ERR(\%)}  & SCLSTM & 2.978          & 1.666          & 4.076          & 2.599          & 2.964          \\
                      & SCVAE  & \textbf{2.823} & \textbf{1.528} & \textbf{2.819} & \textbf{1.841} & \textbf{2.148} \\ \hline
\multirow{2}{*}{BLEU} & SCLSTM & 0.529          & 0.642          & 0.475          & 0.439          & 0.476          \\
                      & SCVAE  & \textbf{0.540} & \textbf{0.652} & \textbf{0.478} & \textbf{0.442} & \textbf{0.478} \\ \hline
\multirow{2}{*}{PPL}  & SCLSTM & 2.654          & 3.229          & 3.365          & 3.941          & 3.556          \\
                      & SCVAE  & \textbf{2.649} & \textbf{3.159} & \textbf{3.337} & \textbf{3.919} & \textbf{3.528} \\ \hline
\end{tabular}%
}
\end{table}
\subsubsection{Cross-domain Training}
Table~\ref{tab:indiv} shows the comparison between SCVAE and SCLSTM. Both are trained on the full cross-domain dataset, and tested on the four domains individually. The SCVAE outperforms the SCLSTM on all metrics. For the highly complex TV and laptop domains, the SCVAE leads to dramatic improvements in ERR. This shows that the additional sentence level conditioning through $z$ helps to convey all desired attributes.


\subsubsection{Limited Training Data}
\begin{figure}[bp]
	\centering
	\includegraphics[width=.875\linewidth]{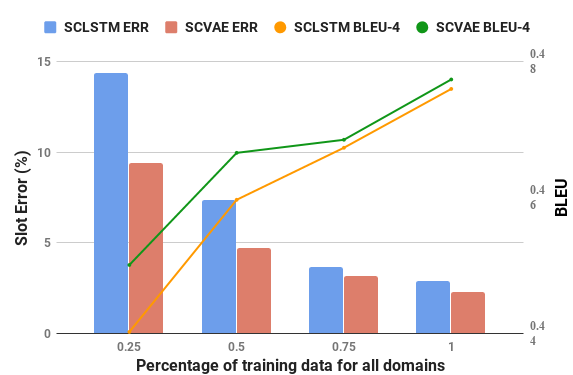}
    \caption{Comparison between SCVAE and SCLSTM with limited training data.}
    \label{fig:percent}
\end{figure}

Fig.~\ref{fig:percent} shows BLEU and ERR results when the SCVAE and SCLSTM are trained on varying amounts of data. The SCVAE has a lower ERR than the SCLSTM across the varying amounts of training data. For very slow amounts of data the SCVAE outperforms the SCLSTM even more. 
In addition, our model consistently achieves better results on the BLEU metric.


\subsubsection{K-Shot Learning}
For the K-shot learning experiments, we trained the model using all training examples from three domains and only 300 examples from the target domain\footnote{600 examples were used for laptop as target domain.}. The target domain is the domain we test on. As seen from Table~\ref{tab:shot}, the SCVAE outperforms the SCLSTM in all domains except hotel. This might be because the hotel domain is the simplest and the model does not need to rely on the knowledge from other domains. The SCVAE strongly outperforms the SCLSTM for the complex TV and laptop domains where the number of distinct SRs is large. This suggests that the SCVAE is better at transferring knowledge between domains.


\begin{table}[tp]
\centering
\caption{Comparison between SCVAE and SCLSTM in K-shot learning}
\label{tab:shot}
\resizebox{\columnwidth}{!}{%
\begin{tabular}{l|l|cccc}
\hline
Metrics                  & Method & Restaurant      & Hotel          & TV              & Laptop          \\ \hline\hline
\multirow{2}{*}{ERR(\%)} & SCLSTM & 13.039          & \textbf{5.366} & 24.497          & 27.587          \\
                         & SCVAE  & \textbf{10.329} & 6.182          & \textbf{20.590} & \textbf{20.864} \\ \hline
\multirow{2}{*}{BLEU}    & SCLSTM & \textbf{0.462}  & 0.578          & 0.382           & 0.379           \\
                         & SCVAE  & 0.458           & \textbf{0.579} & \textbf{0.397}  & \textbf{0.393}  \\ \hline
\multirow{2}{*}{PPL}     & SCLSTM & 3.649           & 4.861          & 5.171           & 6.469           \\
                         & SCVAE  & \textbf{3.575}  & \textbf{4.800} & \textbf{5.092}  & \textbf{6.364}  \\ \hline
\end{tabular}%
}
\end{table}

\section{Conclusion}
In this paper, we propose a semantically conditioned variational autoencoder (SCVAE) for natural language generation. The SCVAE encodes information about both the semantic representation and the sentence into a latent variable $z$. Due to a newly proposed regularization method, the latent variable $z$ contains semantically meaningful information. 
Therefore, conditioning on $z$ leads to a strong improvement in generating sentences with all desired attributes. In an extensive comparison the SCVAE outperforms the SCLSTM on a range of metrics when training on different sizes of data and for K-short learning. Especially, when testing the ability to convey all desired information within complex domains, the SCVAE shows significantly better results.



\section*{Acknowledgments}
Bo-Hsiang Tseng is supported by Cambridge Trust and the Ministry of Education, Taiwan.
This research was partly funded by the EPSRC grant EP/M018946/1 Open Domain Statistical Spoken Dialogue Systems. Florian Kreyssig is supported by the Studienstiftung des Deutschen Volkes. Pawe{\l} Budzianowski is supported by the EPSRC and Toshiba Research Europe Ltd.

\bibliography{acl2018}
\bibliographystyle{acl_natbib}

\newpage
\appendix

\begin{table*}
\centering
\caption{Semantic representation (SR) with ground truth (GT) and sentences generated by SCVAE}
\label{tab:example}
\resizebox{\textwidth}{!}{%
\begin{tabular}{ll}
\hline
\multicolumn{2}{c}{\textbf{Restaurant Domain}}                                                                                                                                                                                \\ \hline\hline
SR   & inform(name='la mediterranee';food='middle eastern';kidsallowed=no;pricerange=cheap)                                                                                                                                   \\
GT   & i have just the restaurant for you . it is called la mediterranee , it serves cheap middle eastern food and childs are not allowed                                                                                     \\
Gen1 & la mediterranee serves middle eastern food in the cheap price range and does not allow childs                                                                                                                          \\
Gen2 & la mediterranee is a cheap middle eastern restaurant that does not allow kids                                                                                                                                          \\
Gen3 & la mediterranee is cheaply priced restaurant serves middle eastern food and allow childs                                                                                                                               \\
Gen4 & i would recommend la mediterranee . it is cheap middle eastern food , does not allow child                                                                                                                             \\
Gen5 & la mediterranee does not allow kids , serves middle eastern food and it is cheap price                                                                                                                                 \\ \hline
\multicolumn{2}{c}{\textbf{Hotel Domain}}                                                                                                                                                                                     \\ \hline\hline
SR   & inform\_count(type='hotel';count='2';near='marina cow hollow';pricerange='inexpensive')                                                                                                                                \\
GT   & there are 2 other hotels near marina cow hollow that fit inexpensive your price range                                                                                                                                  \\
Gen1 & there are 2 inexpensive hotels near marina cow hollow                                                                                                                                                                  \\
Gen2 & there are 2 hotels near the marina cow hollow that are inexpensive                                                                                                                                                     \\
Gen3 & there are 2 inexpensively priced hotel near marina cow hollow                                                                                                                                                          \\
Gen4 & there are 2 inexpensive priced hotels near the marina cow hollow                                                                                                                                                       \\
Gen5 & there are 2 hotels in the inexpensive price range near marina cow hollow                                                                                                                                               \\ \hline
\multicolumn{2}{c}{\textbf{Television Domain}}                                                                                                                                                                                     \\ \hline\hline
SR   & recommend(name=hymenaios 11;type=television;family=d1;hdmiport=1)                                                                                                                                                      \\
GT   & the hymenaios 11 is a television with 1 hdmi port in the d1 product family                                                                                                                                             \\
Gen1 & the name of hymenaios 11 is a television in the d1 family with 1 hdmi port                                                                                                                                             \\
Gen2 & hymenaios 11 is a television from the d1 product family with 1 hdmi port                                                                                                                                               \\
Gen3 & the hymenaios 11 television is a member of the d1 product family and has 1 hdmi port                                                                                                                                   \\
Gen4 & the hymenaios 11 television is part of the d1 family , and has 1 hdmi port                                                                                                                                             \\
Gen5 & the hymenaios 11 is a nice television in the d1 family with 1 hdmi port                                                                                                                                                \\ \hline
\multicolumn{2}{c}{\textbf{Laptop Domain}}                                                                                                                                                                                    \\ \hline\hline
SR   & inform\_no\_match(type=laptop;isforbusinesscomputing=true;driverange=medium;weightrange=mid weight)                                                                                                                    \\
GT   & no matches were found for a laptop search for business computing with a medium drive that falls in the mid weight range                                                                                                \\
Gen1 & there is no laptop in the medium drive range that is mid weight range and are used for business use                                                                                                                    \\
Gen2 & there are no laptops in the medium drive size range and is in the mid weight range . they are used for business computing                                                                                              \\
Gen3 & sorry but there are no laptops that match for a medium drive range , mid weight weight range , and is for business computing                                                                                           \\
Gen4 & there are no mid weight laptops with medium driver and are used for business computing                                                                                                                                 \\
Gen5 & \begin{tabular}[c]{@{}l@{}}unfortunately , we have no matches for your requirementss for a laptop in the mid weight range , that can be used for \\ business computing , and is in the medium drive range\end{tabular}
\end{tabular}%
}
\end{table*}

\end{document}